%% file: eccv2018submission.tex
\documentclass[runningheads]{llncs}
\usepackage{hyperref}
\hypersetup{
    colorlinks=true,
    linkcolor=blue,
    filecolor=magenta,      
    urlcolor=magenta,
}
\usepackage{graphicx}
\usepackage{mathtools}
\usepackage{booktabs}
\usepackage{makecell}
\usepackage{multirow}
\usepackage{cite}
\usepackage{siunitx}
\usepackage{amsmath,amssymb} 
\usepackage[usenames, dvipsnames]{color}
\usepackage{pifont}
\newcommand{\cmark}{\color{OliveGreen}\ding{51}}%
\newcommand{\xmark}{\color{BrickRed}\ding{55}}%
\usepackage[width=122mm,left=12mm,paperwidth=146mm,height=193mm,top=12mm,paperheight=217mm]{geometry}

\newcommand{\etal}{\textit{et al}.}

\newcommand{\dataset}{\mbox{\sc Flintstones}}
\newcommand{\taskfull}{\mbox{Semantic Scene Generation}}
\newcommand{\task}{\mbox{SSG}}
\newcommand{\modelfull}{\mbox{\textbf{C}omposition, \textbf{R}etrieval \textbf{a}nd \textbf{F}usion Ne\textbf{t}work}}
\newcommand{\model}{\mbox{\sc Craft}}

\begin{document}
\pagestyle{headings}
\mainmatter
\def\ECCV18SubNumber{2869}  

\title{Imagine This! Scripts to Compositions to Videos}





\author{
Tanmay Gupta\textsuperscript{1}
\and Dustin Schwenk\textsuperscript{2} 
\and Ali Farhadi\textsuperscript{2,3} 
\and \\ Derek Hoiem\textsuperscript{1} 
\and Aniruddha Kembhavi\textsuperscript{2}}
\institute{
University of Illinois Urbana-Champaign
\and 
Allen Institute for Artificial Intelligence 
\and
University of Washington}

\maketitle
\vspace{-2.0em}
\begin{abstract}

Imagining a scene described in natural language with realistic layout and appearance of entities is the ultimate test of spatial, visual, and semantic world knowledge. Towards this goal, we present the \modelfull\ (\model), a model capable of learning this knowledge from video-caption data and applying it while generating videos from novel captions. \model\ explicitly predicts a temporal-layout of mentioned entities (characters and objects), retrieves spatio-temporal entity segments from a video database and fuses them to generate scene videos. Our contributions include sequential training of components of \model\ while \emph{jointly} modeling layout and appearances, and losses that encourage learning compositional representations for retrieval. We evaluate \model\ on \emph{semantic fidelity} to caption, \emph{composition consistency}, and \emph{visual quality}. \model\ outperforms direct pixel generation approaches and generalizes well to unseen captions and to unseen video databases with no text annotations. We demonstrate \model\ on \dataset, a new richly annotated video-caption dataset with over 25000 videos. For a glimpse of videos generated by \model, see \url{https://youtu.be/688Vv86n0z8}.

\input{figures/hero.tex}

\end{abstract}

\input{sections/01_introduction.tex}
\input{sections/02_related_work.tex}
\input{sections/03_model.tex}
\input{sections/04_dataset.tex}

\input{sections/05_experiments.tex}

\bibliographystyle{splncs}
\bibliography{00_references}
\clearpage
\appendix
\input{sections/appendix.tex}

\end{document}

%% file: figures/hero.tex
\begin{figure}[ht]
\centering
\vspace{-1.5em}
\includegraphics[width=\textwidth]{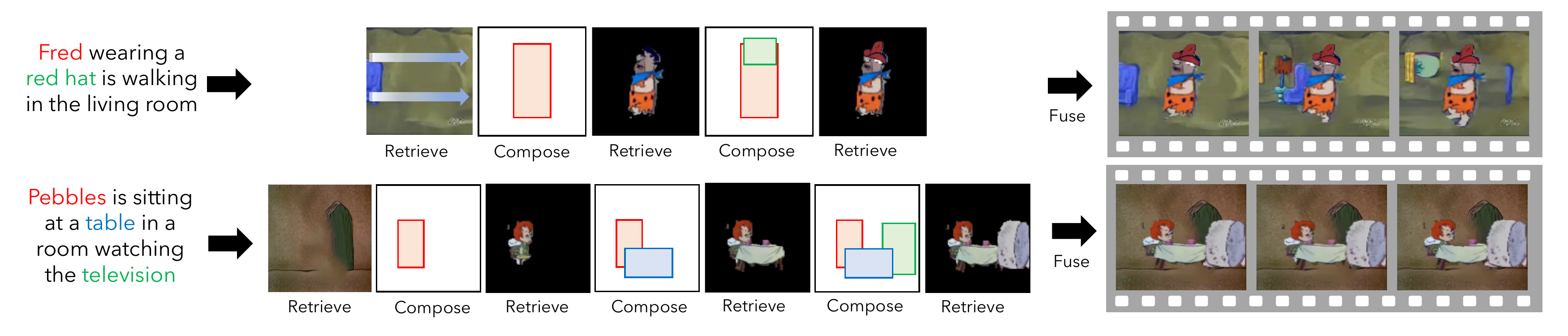}
\vspace{-2em}
\caption{Given a novel description, \model\ sequentially composes a scene layout and retrieves entities from a video database to create complex scene videos.}
\vspace{-4.0em}
\label{fig:hero}
\end{figure}

%% file: sections/01_introduction.tex
\section{Introduction}
\vspace{-0.6em}
Consider the scene description: \emph{Fred is wearing a blue hat and talking to Wilma in the living room. Wilma then sits down on a couch}. Picturing the scene in our mind requires the knowledge of plausible locations, appearances, actions, and interactions of characters and objects being described, as well as an ability to understand and translate the natural language description into a plausible visual instantiation. In this work, we introduce \taskfull\ (\task), the task of generating complex scene videos from rich natural language descriptions which requires jointly modeling the layout and appearances of entities mentioned in the description. \task\ models are trained using a densely annotated video dataset with scene descriptions and entity bounding boxes. During inference, the models must generate videos for novel descriptions (unseen during training).

Modelling the layout and appearances of entities for descriptions like the one above poses several challenges: (a) \textbf{Entity Recall} - the video must contain the relevant characters (Fred, Wilma), objects (blue hat, couch) and background (setting that resembles a living room); (b) \textbf{Layout Feasibility} - characters and objects must be placed at plausible locations and scales (Fred, Wilma and the couch should be placed on the ground plane, the hat must lie on top of Fred's head); (c) \textbf{Appearance Fidelity} - entity appearance, which may be affected by identity, pose, action, attributes and layout, should respect the scene description; (d) \textbf{Interaction Consistency} - appearance of characters and objects must be \emph{consistent with each other} given the described, sometimes implicit, interaction (Fred and Wilma should face each other as do people when they talk to each other); (f) \textbf{Language Understanding} - the system must be able to understand and translate a natural language description into a plausible visual instantiation.

Currently, the dominant approaches to conditional generation of visual data from text rely on directly learning distributions in a \emph{high dimensional pixel space}. While these approaches have shown impressive results for aligned images of objects (faces, birds, flowers, etc.), they are often inadequate for addressing the above challenges, due to the \emph{combinatorial explosion} of the image space arising from multiple characters and objects with significant appearance variations arranged in a large number of possible layouts. In contrast, our proposed \modelfull\ (\model) explicitly models the spatio-temporal layout of characters and objects in the scene jointly with entity appearances. Unlike pixel generation approaches, our appearance model is based on text to entity segment retrieval from a video database. Spatio-temporal segments are extracted from the retrieved videos and fused together to generate the final video. The layout composition and entity retrieval work in a sequential manner which is determined by the language input. Factorization of our model into composition and retrieval stages alleviates the need to directly model pixel spaces, results in an architecture that more easily exploits location and appearance contextual cues, and renders a more interpretable output.

Towards the goal of \task, we introduce \dataset, a densely annotated dataset based on \emph{The Flintstones} animated series, consisting of over 25000 videos, each 75 frames long. \dataset\ has several advantages over using a random sample of internet videos. First, in a closed world setting such as a television series, the most frequent characters are present in a wide variety of settings, which serves as a more manageable learning problem than a sparse set obtained in an open world setting. Second, the flat textures in animations are easier to model than real world videos. Third, in comparison to other animated series, The Flintstones has a good balance between having fairly complex interactions between characters and objects while not having overly complicated, cluttered scenes. For these reasons, we believe that the \dataset\ dataset is semantically rich, preserves all the challenges of text to scene generation and is a good stepping stone towards real videos. \dataset\ consists of an $80$-$10$-$10$ train-val-test split. The train and val sets are used for learning and model selection respectively. Test captions serve as novel descriptions to generate videos at test time. To quantitatively evaluate our model, we use two sets of metrics. The first measures \emph{semantic fidelity} of the generated video to the desired description using entity noun, adjective, and verb recalls. The second measures \emph{composition consistency}, \emph{i.e.} the consistency of the appearances, poses and layouts of entities with respect to other entities in the video and the background.

We use \dataset\ to evaluate \model\ and provide a detailed ablation analysis. \model\ outperforms baselines that generate pixels directly from captions as well as a whole video retrieval approach (as opposed to modeling entities). It generalizes well to unseen captions as well as unseen videos in the target database. Our quantitative and qualitative results show that for simpler descriptions, \model\ exploits location and appearance contextual cues and outputs videos that have consistent layouts and appearances of described entities. However, there is tremendous scope for improvement. \model\ can fail catastrophically for complex descriptions (containing large number of entities, specially infrequent ones). The adjective and verb recalls are also fairly low. We believe \task\ on \dataset\ presents a challenging problem for future research.


%% file: sections/02_related_work.tex
\vspace{-1.0em}
\section{Related Work}
\vspace{-0.6em}
\noindent \textbf{Generative models} Following pioneering work on Variational Autoencoders \cite{Kingma2013AutoEncodingVB} and Generative Adversarial Networks \cite{Goodfellow2014GenerativeAN}, there has been tremendous interest in generative modelling of visual data in a high dimensional pixel space. Early approaches focused on unconditional generation~\cite{Bengio2013BetterMV,Arjovsky2017WassersteinG,Radford2015UnsupervisedRL,Gregor2015DRAWAR}, whereas recent works have explored models conditioned on simple textual inputs describing objects~\cite{Reed2016LearningWA,Reed2016GenerativeAT,Zhang2016StackGANTT,mansimov16_text2image,Yan2016Attribute2ImageCI}. While the visual quality of images generated by these models has been steadily improving~\cite{Odena2017ConditionalIS,Karras2017ProgressiveGO}, success stories have been limited to generating images of aligned objects (e.g. faces, birds, flowers), often training one model per object class.  In contrast, our work deals with generating complex scenes which requires modelling the layout and appearances of multiple entities in the scene. 

Of particular relevance is the work by Hong~\etal~\cite{Hong2018InferringSL} who first generate a coarse semantic layout of bounding boxes, refine that to segmentation masks and then generate an image using an image-to-image translation model~\cite{Isola2016ImagetoImageTW,Chen2017PhotographicIS}. A limitation of this approach is that it assumes a fixed number of object classes (80 in their experiments) and struggles with the usual challenge of modeling high dimensional pixel spaces such as generating coherent entities. Formulating appearance generation in terms of entity retrieval from a database allows our model to scale to a large number of entity categories, guarantee intra-entity coherence and allows us to focus on the semantic aspects of scene generation and inter-entity consistency. The retrieval approach also lends itself to generating videos without significant modification. There have been attempts at extending GANs for unconditional ~\cite{Vondrick2016GeneratingVW,villegas2017HierPred} as well as text conditional~\cite{Marwah2017AttentiveSV} video generation, but quality of generated videos is usually worse than that of GAN generated images unless used in very restrictive settings. A relevant generative modelling approach is by Kwak~\etal~\cite{Kwak2016GeneratingIP} who proposed a model in which parts of the image are generated sequentially and combined using alpha blending. However, this work does not condition on text and has not been demonstrated on complex scenes. Another relevant body of work is by Zitnick~\etal~\cite{Zitnick2013BringingSI,Zitnick2013LearningTV,Zitnick2016AdoptingAI} who compose static images from descriptions with clipart images using a Conditional Random Field formulation.

To control the structure of the output image, a growing body of literature conditions image generation on a wide variety of inputs ranging from keypoints~\cite{Reed2017GeneratingII} and sketches~\cite{Liu2017AutopainterCI} to semantic segmentation maps~\cite{Isola2016ImagetoImageTW}. In contrast to these approaches which condition on \emph{provided} location, our model \emph{generates} a plausible scene layout and then conditions entity retrieval on this layout.

\noindent \textbf{Phrase Grounding and Caption-Image Retrieval.} The entity retriever in \model\ is related to caption based image retrieval models. The caption-image embedding space is typically learned by minimizing a ranking loss such as a triplet loss~\cite{Frome2013Devise,Kiros2014UnifyingVSE,Frome2013Devise,Wang2016StructPreservEmbed,Faghri2017VSE++}. Phrase grounding~\cite{Plummer2015Flickr30k} is another closely related task where the goal is to localize a region in an image described by a phrase. 

One of our contributions is enriching the semantics of embeddings learned through triplet loss by simultaneously minimizing an auxiliary classification loss based on noun, adjective and verb words associated with an entity in the text description. This is similar in principle to \cite{Tsai2017RobustVSE} where auxiliary autoencoding losses were used in addition to a primary binary prediction loss to learn robust visual semantic embeddings. Learning shared representations across multiple related tasks is a key concept in multitask learning~\cite{Caruana1998Multitask,Gupta2017Aligned}.

%% file: sections/03_model.tex
\vspace{-0.9em}
\section{Model}
\vspace{-0.6em}
\input{figures/overview.tex}
Figure~\ref{fig:overview} presents an overview of \modelfull\ which consists of three parts: \emph{Layout Composer}, \emph{Entity Retriever}, and \emph{Background Retriever}. Each is a neural network that is trained independently using ground truth supervision. At inference time, \model\ begins with an empty video and adds entities in the scene sequentially based on the order of appearance in the description. At each step, the \emph{Layout Composer} predicts a location and scale for an entity given the text and the video constructed so far. Then, conditioned on the predicted location, text, and the partially constructed video, the \emph{Entity Retriever} produces a query embedding that is looked up against the embeddings of entities in the target video database. The entity is cropped from the retrieved video and placed at the predicted location and scale in the video being generated. Alternating between the \emph{Layout Composer} and \emph{Entity Retriever} allows the model to condition the layout of entities on the appearance and vice versa. Similar to \emph{Entity Retriever}, the \emph{Background Retriever} produces a query embedding for the desired scene from text and retrieves the closest background video from the target database. The retrieved spatio-temporal entity segments and background are fused to generate the final video. We now present the notation used in the rest of the paper, followed by architecture and training details for the three components.

\begin{center}
\vspace{-0.4em}
\begingroup
\footnotesize
\noindent\begin{tabular}{@{}|l | l|} 
  \hline
  \textbf{Caption} & \\
  $T$ & Caption with length $|T|$\\
  $\{E_i\}_{i=1}^{n}$ & $n$ entities in $T$ in order of appearance\\ 
  $\{e_i\}_{i=1}^{n}$ & entity noun positions in $T$\\
  \hline
  \textbf{Video} & \\
  $F$ & number of frames in a video\\
  $\{(l_i,s_i)\}_{i=1}^{n}$ & position of entities in the video\\
  $l_i$ & entity bounding box at each frame ($\{(x_{if},y_{if},w_{if},h_{if})\}_{f=1}^{F}$)\\
  $s_i$ & entity pixel segmentation mask at each frame\\
  $V_{i-1}$ & partially constructed video with entities $\{E_j\}_{j=1}^{i-1}$\\
  $V$ ($=V_n$) & full video containing all entities\\
  \hline
  $\{(V^{[m]},T^{[m]})\}_{m=1}^{M}$ & training data points, where $M=$ number of data points\\
  \hline
\end{tabular}
\vspace{-0.75em}
\endgroup
\end{center}



\subsection{Layout Composer}

The layout composer is responsible for generating a plausible layout of the scene consisting of the locations and scales of each character and object mentioned in the scene description. Jointly modeling the locations of all entities in a scene presents fundamentally unique challenges for spatial knowledge representation beyond existing language-guided localization tasks. Predicting plausible locations and scales for objects not yet in an image requires a significant amount of \emph{spatial knowledge} about people and objects, in contrast to text based object localization which relies heavily on appearance cues. This includes knowledge like -- a hat goes on top of a person's head, a couch goes under the person sitting on it, a person being talked to faces the person speaking instead of facing away, tables are short and wide while standing people are tall and thin, etc.

Figure~\ref{fig:layout_predictor} presents a schematic for the layout composer. Given the varying number of entities across videos, the layout composer is setup to run in a sequential manner over the set of distinct entities mentioned in a given description. At each step, a text embedding of the desired entity along with a partially constructed video (consisting of entities fused into the video at previous steps) are input to the model which predicts distributions for the location and scale of the desired entity. 

\input{figures/layout_predictor.tex}

The layout composer models $P(l_i|V_{i-1},T,e_i;\theta_{loc},\theta_{sc})$, the conditional distribution of the location and scale (width and height normalized by image size) of the $i^{th}$ entity given the text, entity noun position in tokenized text, and the partial video with previous entities. Let $C_i$ denote the conditioning information, $(V_{i-1},T,e_i)$. We factorize the position distribution into location and scale components as follows:
\begin{small}
\vspace{-1.0em}
\begin{equation}
P(l_i|C_i;\theta_{loc},\theta_{sc}) = \prod_{f=1}^{F} P_{loc}^f(x_{if},y_{if}|C_i;\theta_{loc}^f)\cdot P_{sc}^f(w_{if},h_{if}|x_{if},y_{if},C_i;\theta_{sc}^f)
\end{equation}
\vspace{-1.0em}
\end{small}

$\theta_{loc}=\{\theta_{loc}^f\}_{f=1}^{F}$ and $\theta_{sc}=\{\theta_{sc}^f\}_{f=1}^{F}$ are learnable parameters. $P_{loc}^f$ is modelled using a network that takes $C_i$ as input and produces a distribution over all pixel locations for the $f^{th}$ image frame. We model $P_{sc}^f$ using a Gaussian distribution whose mean $\mu_f$ and covariance $\Sigma_f$ are predicted by a network given 
$(x_i,y_i,C_i)$. Parameters $\theta_{loc}$ and $\theta_{sc}$ are learned from ground truth position annotations by minimizing the following maximum likelihood estimation loss:


\begin{small}
\vspace{-1.0em}
\begin{multline}\label{eq:mle_loss}
\sum_{m=1}^{M}\sum_{i=1}^{n^{[m]}}\sum_{f=1}^{F}\left[-\log(P_{loc}^f(x_{if}^{[m]},y_{if}^{[m]}|C_i^{[m]};\theta_{loc}^f)) + 0.5 \cdot  \log(\det(\Sigma(x_{if},y_{if},C_i;\theta_{sc}^f))){+}\right. \\
\left. 0.5 \cdot (z_{if}^{[m]}-\mu_f(D_{i}^{[m]};\theta_{sc}^f))^T{\Sigma_f}^{-1}(z_{if}^{[m]}-\mu_f(D_{i}^{[m]};\theta_{sc}^f)) + \log(2\pi)
\right]
\vspace{-1.0em}
\end{multline}
\end{small}
\noindent where \begin{footnotesize}$z_{if}=[w_{if};h_{if}]$ \& $D_{i}^{[m]}=(x_{i}^{[m]},y_{i}^{[m]},C_i^{[m]})$\end{footnotesize}. For simplicity, we manually set and freeze $\Sigma$ to an isometric diagonal covariance matrix  with variance of $0.005$.\\

\vspace{-0.6em}
\noindent \textbf{Feature Computation Backbone.} The location and scale predictors have an identical feature computation backbone comprising of a CNN and a bidirectional LSTM. The CNN encodes $V_{i-1}$ (8 sub-sampled frames concatenated along the channel dimension) as a set of convolutional feature maps which capture appearance and positions of previous entities in the scene. The LSTM is used to encode the entity $E_i$ for which the prediction is to be made along with semantic context available in the caption. The caption is fed into the LSTM and the hidden output at $e_i^{th}$ word position is extracted as the entity text encoding. The text encoding is replicated spatially and concatenated with convolutional features and 2-D grid coordinates to create a representation for each location in the convolutional feature grid that is aware of visual, spatial, temporal, and semantic context.\\

\noindent \textbf{Location Predictor.} $P_{loc}^f$ is modelled using a Multi Layer Perceptron (MLP) that produces a score for each location. This map is bilinearly upsampled to the size of input video frames. Then, a softmax layer over all pixels produces $P_{loc}^f(x,y|C;\theta_{loc}^f)$ for every pixel location $(x,y)$ in the $f^{th}$ video frame.\\

\noindent \textbf{Scale Predictor.} Features computed by the backbone at a particular $(x,y)$ location are selected and fed into the scale MLP that produces $\mu_f(x_i,y_i,C_i;\theta_{sc}^f)$.\\

\noindent \textbf{Feature sharing and multitask training.} While it is possible to train a separate network for each $\{P_{loc}^f,\mu_f\}_{f=1}^F$, we present a pragmatic way of sharing features and computation for different frames and also between the location and scale networks. To share features and computation across frames, the location network produces $F$ probability maps in a single forward pass. This is equivalent to sharing all layers across all $P_{loc}^f$ nets except for the last layer of the MLP that produces location scores. Similarly, all the $\mu_f$ nets are also combined into a single network. We refer to the combined networks by $P_{loc}$ and $\mu$.

In addition, we also share features across the location and scale networks. First, we share the feature computation backbone, the output from which is then passed into location and scale specific layers. Second, we use a soft-attention mechanism to select likely positions for feeding into the scale layers. This conditions the scale prediction on the plausible locations of the entity. We combine the $F$ spatial maps into a single attention map through max pooling. This attention map is used to perform weighted average pooling on backbone features and then fed into the scale MLP. Note that this is a differentiable greedy approximation to find the most likely location (by taking argmax of spatial probability maps) and scale (directly using output of $\mu$, the mode for a gaussian distribution) in a single forward pass. To keep training consistent with inference, we use the soft-attention mechanism instead of feeding ground-truth locations into $\mu$. 

\vspace{-1.0em}
\subsection{Entity Retriever}
\vspace{-0.2em}
The task of the entity retriever is to find a spatio-temporal patch within a target database that matches an entity in the description \emph{and} is consistent with the video constructed thus far -- the video with all previous entities retrieved and placed in the locations predicted by the layout network. We adopt an embedding based lookup approach for entity retrieval. This presents several challenges beyond traditional image retrieval tasks. Not only does the retrieved entity need to match the semantics of the description but it also needs to respect the implicit relational constraints or context imposed by the appearance and locations of other entities. E.g. for \emph{Fred is talking to Wilma}, it is not sufficient to retrieve \emph{a Wilma}, but one who is also facing in the right direction, i.e. towards \emph{Fred}.

\input{figures/entity_retriever.tex}
The Entity Retriever is shown in Figure~\ref{fig:entity_retriever} and consists of two parts: (i) query embedding network $Q$, and (ii) target embedding network $R$. $Q$ and $R$ are learned using the query-target pairs \begin{footnotesize}$\big \langle (T^{[m]},e_i^{[m]},l_i^{[m]},V_{i-1}^{[m]}),(V^{[m]},l_i^{[m]},s_i^{[m]})\rangle_{i,m}$\end{footnotesize} in the training data. For clarity, we abbreviate \begin{footnotesize}$Q(T^{[m]},e_i^{[m]},l_i^{[m]},V_{i-1}^{[m]})$\end{footnotesize} as $q_i^{[m]}$ and \begin{footnotesize}$R(V^{[m]},l_i^{[m]},s_i^{[m]})$  as $r_i^{[m]}$\end{footnotesize}. At each training iteration, we sample a mini-batch of $B$ pairs without replacement and compute embeddings \begin{footnotesize}$\{(q_{i_b}^{[m_b]},r_{i_b}^{[m_b]})\}_{b=1}^{B}$\end{footnotesize} where $q$ and $r$ are each sequence of $F$ embeddings corresponding to $F$ video frames. The model is trained using a triplet loss computed on \emph{all possible} triplets in the mini-batch. Let $\delta_b$ denote the set of all indices from $1$ to $B$ except $b$. The loss can then be defined as 

\begin{footnotesize}
\vspace{-2.0em}
\begin{multline}
\mathcal{L}_{triplet} = \frac{1}{B\cdot(B-1)}\sum_{b=1}^{B}\sum_{b^-\in \delta_b} \left[ \max(0,\gamma + q_{i_b}^{[m_b]}\odot r_{i_{b^-}}^{[m_{b^-}]} - q_{i_b}^{[m_b]}\odot r_{i_{b}}^{[m_{b}]}) \; + \right.\\
\left. \max(0,\gamma + q_{i_{b^-}}^{[m_{b^-}]}\odot r_{i_b}^{[m_b]} - q_{i_b}^{[m_b]}\odot r_{i_{b}}^{[m_{b}]})
\right]
\vspace{-1.5em}
\end{multline}
\end{footnotesize}
\noindent where $q\odot r=\frac{1}{F}\sum_{f=1}^{F}q[f] \cdot r[f]$ is the average dot product between corresponding query and target frame embeddings. We use a margin of $\gamma=0.1$.\\

\vspace{-0.7em}
\noindent \textbf{Auxiliary Multi-label Classification Loss} 
We found that models trained using triplet loss alone could simply learn a one-to-one mapping between ground truth text and entity video segments with poor generalization to unseen captions and database videos. To guide the learning to utilize the compositional nature of text and improve generalization, we add an auxiliary classification loss on top of the embeddings. The key idea is to enrich the semantics of the embedding vectors by predicting the noun, adjectives, and action words directly associated with the entity in the description. For example, \emph{Wilma}'s embedding produced by the query and target embedding networks in \emph{Fred is talking to a happy Wilma who is sitting on a chair.} is forced to predict \emph{Wilma}, \emph{happy} and \emph{sitting} ensuring their representation in the embeddings. A vocabulary $\mathcal{W}$ is constructed of all nouns, adjectives and verbs appearing in the training data. Then for each sample in the mini-batch, an MLP is used as a multi-label  classifier to predict associated words from the query and target embeddings. Note that a single MLP is used to make these noun, adjective and verb predictions on \emph{both} query and target embeddings. \\


\vspace{-0.7em}
\noindent \textbf{Query Embedding Network ($Q$).} Similar to the layout composer's feature computation backbone, $Q$ consists of a CNN to independently encode every frame of $V_{i-1}$ and an LSTM to encode $(T,e_i)$ which are concatenated together along with a 2-D coordinate grid to get per-frame feature maps. However, unlike layout composer, the query embedding network also needs to be conditioned on the position $l_i$ where entity $E_i$ is to be inserted in $V_{i-1}$. To get location and scale specific query embeddings, we use a simplified RoIAlign (RoIPool with RoI quantization and bilinear interpolation) mechanism to crop out the per-frame feature maps using the corresponding bounding box $l_i^f$ and scaling it to a $7\times7$ receptive field. The RoIAlign features are then averaged along the spatial dimensions to get the vector representations for each time step independently. An LSTM applied over the sequence of these embeddings is used to capture temporal context. The hidden output of the LSTM at each time step is normalized and used as the frame query embedding $q[f]$. \\

\vspace{-0.5em}
\noindent \textbf{Target Embedding Network ($R$).} Since during inference, $R$ needs to embed entities in the target database which do not have text annotations, it does not use $T$ as an input. Thus, $R$ is similar to $Q$ but without the LSTM to encode the text. In our experiments we found that using 2-D coordinate features in both query and target networks made the network susceptible to ignoring all other features since it provides an easy signal for matching ground truth query-target pairs during training. This in turn leads to poor generalization. Thus, $R$ has no 2-D coordinate features.



\vspace{-0.5em}
\subsection{Background Retriever}
\vspace{-0.5em}
The task of the background retriever is to find a background scene that matches the setting described in the description. To construct a database of backgrounds without characters in them, we remove characters from videos (given bounding boxes) and perform hole filling using PatchMatch~\cite{Barnes2009PatchMatch}. The background retriever model is similar to the entity retriever with two main differences. First, since the whole background scene is retrieved instead of entity segments, the conditioning on position is removed from both query and database embedding networks replacing RoI pooling with global average pooling. Second, while ideally we would like scene and entity retrieval to be conditioned on each other, for simplicity we leave this to future work and currently treat them independently. These modifications essentially reduce the query embedding network to a text Bi-LSTM whose output at the background word location in the description is used as the query embedding, and the target embedding network to a video Bi-LSTM without RoI pooling. The model is trained using just the triplet loss.

%% file: figures/overview.tex
\begin{figure}[t]
\vspace{-1.0em}
\centering
\includegraphics[width=\textwidth]{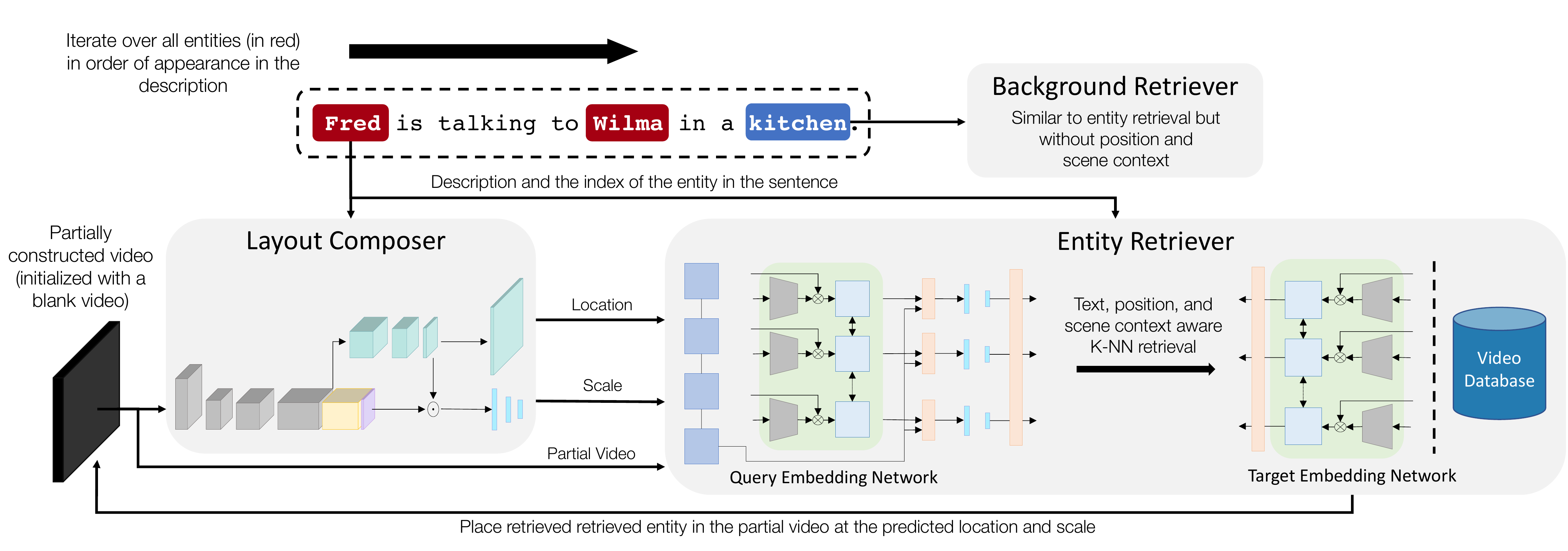}
\vspace{-2.0em}
\caption{\textbf{Overview} of \modelfull\ (\model), consisting of three parts: \emph{Layout Composer}, \emph{Entity Retriever} and \emph{Background Retriever}. \model\ begins with an empty video and sequentially adds entities mentioned in the input description at locations and scales predicted by the Layout Composer.}
\vspace{-1.5em}
\label{fig:overview}
\end{figure}

%% file: figures/layout_predictor.tex
\begin{figure}[t]
\vspace{-1.5em}
\centering
\includegraphics[height=6.5cm]{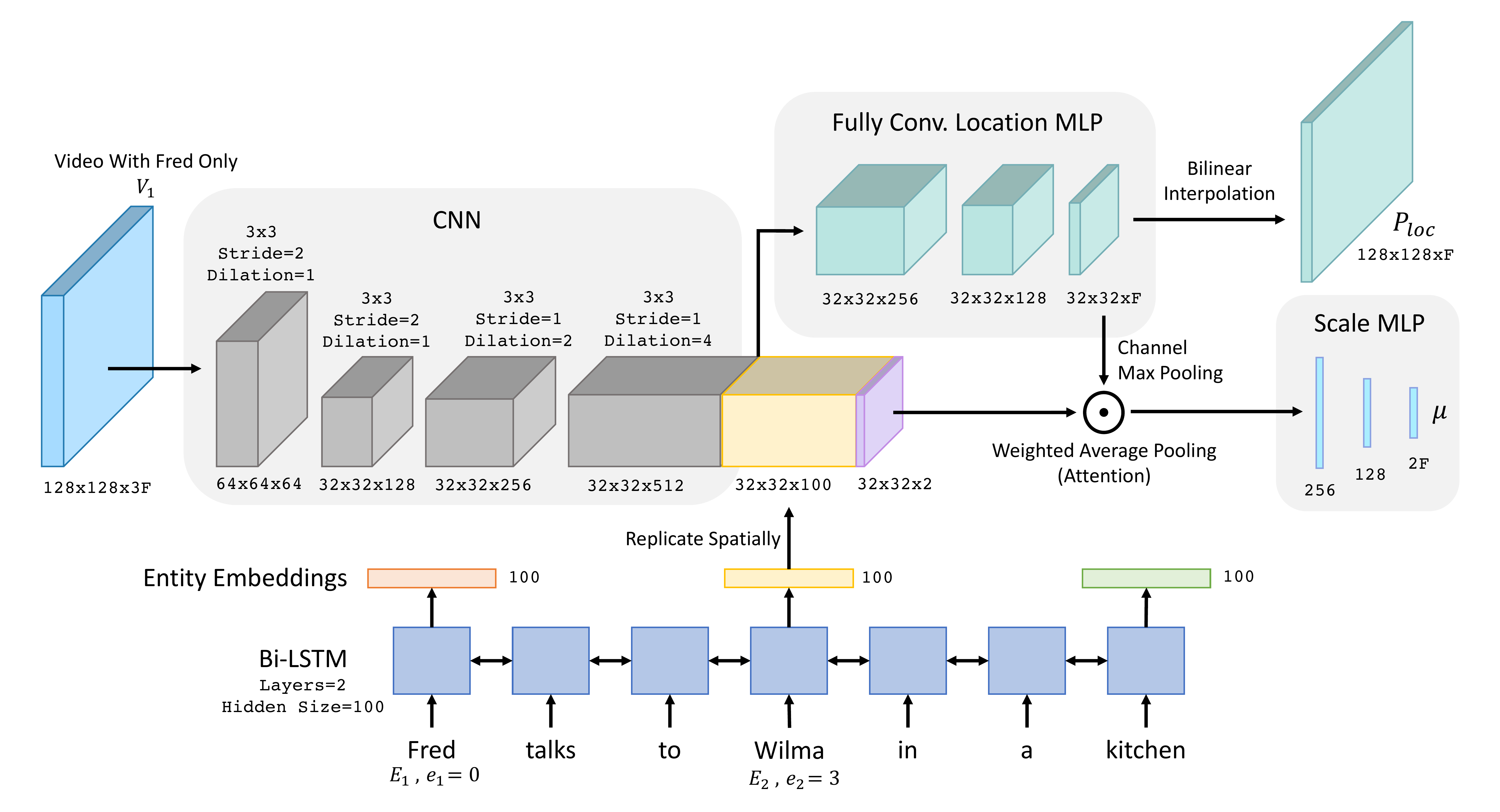}
\vspace{-1.5em}
\caption{\textbf{Layout Composer} is run sequentially through the set of entities in the description, predicting the distributions for the location and scale of the desired entity.}
\vspace{-1.5em}
\label{fig:layout_predictor}
\end{figure}

%% file: figures/entity_retriever.tex
\begin{figure}[t]
\vspace{-0.5em}
\centering
\includegraphics[width=\textwidth]{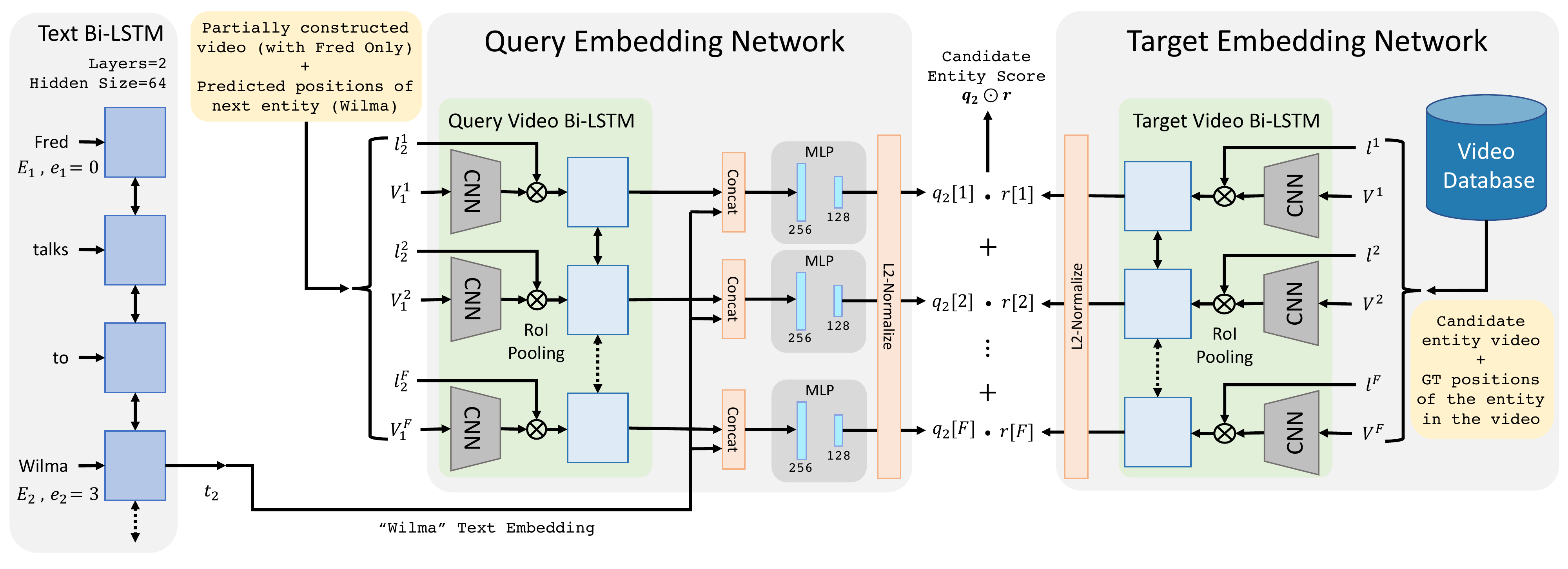}
\vspace{-2.5em}
\caption{\textbf{Entity Retriever} retrieves spatio-temporal patches from a target database that match entity description as encoded by the query embedding network.}
\vspace{-2.0em}
\label{fig:entity_retriever}
\end{figure}

%% file: sections/04_dataset.tex
\section{The Flintstones Dataset}
\vspace{-0.5em}
\input{figures/dataset.tex}

\noindent \textbf{Composition.} The \dataset\ dataset is composed of 25184 densely annotated video clips derived from the animated sitcom \textit{The Flintstones}. Clips are chosen to be 3 seconds (75 frames) long to capture relatively small action sequences, limit the number of sentences needed to describe them and avoid scene and shot changes. As shown in Figure~\ref{fig:dataset_point}, annotations contain clip's characters, setting, and objects being interacted with marked in text as well as their bounding boxes in all frames. \dataset\ has a 80-10-10 train-val-test split. \\

\noindent \textbf{Clip Annotation.} Dense annotations are obtained in a multi-step process: identification and localization of characters in keyframes, identification of the scene setting, scene captioning, object annotation, and entity tracking to provide annotations for all frames. The dataset also contains segmentation masks for characters and objects. First, a rough segmentation mask is produced by using SLIC~\cite{Achanta2012Slic} followed by hierarchical merging. This mask is then used to initialize GrabCut~\cite{Rother2004Grabcut}, which further refines the segmentation. The dataset also contains a clean background for each clip. Foreground characters and objects are excised, and the resulting holes are filled using PatchMatch~\cite{Barnes2009PatchMatch}. \noindent See supplementary material for more details on the dataset.

%% file: figures/dataset.tex
\begin{figure}[t]
\centering
\vspace{-0.5em}
\includegraphics[width=\textwidth]{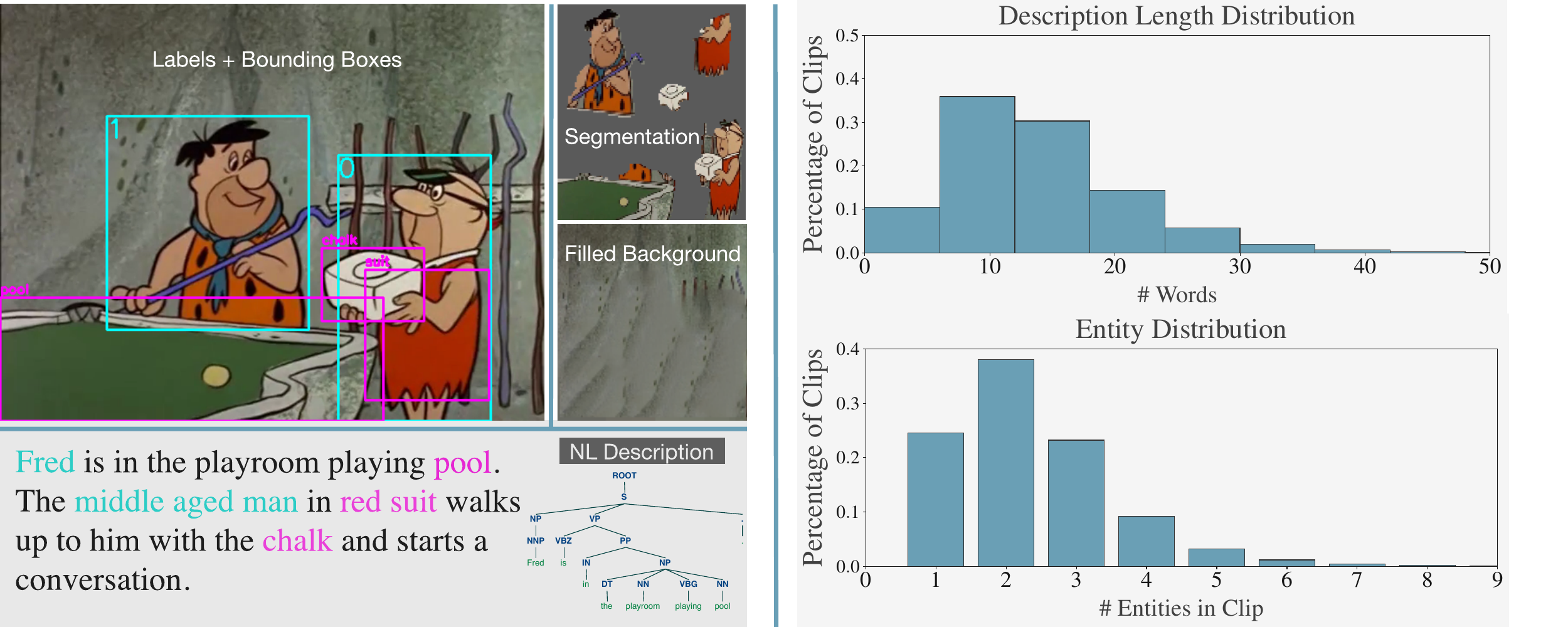}
\vspace{-2.0em}
\caption{\textbf{\dataset\ dataset.} Left: Overview of clip annotations. Right: Distribution of description lengths and entity counts in the dataset.}
\vspace{-1.5em}
\label{fig:dataset_point}
\end{figure}


%% file: sections/05_experiments.tex
\vspace{-1.0em}
\section{Experiments}

\subsection{Layout Composer Evaluation}

\noindent \textbf{Training.} We use the Adam optimizer (learning rate=$0.001$, decay factor=$0.5$ per epoch, weight decay=$0.0001$) and a batch size of 32. \\

\noindent \textbf{Metrics.} We evaluate layout composer using 2 metrics: (a) negative log-likelihood (NLL) of ground truth (GT) entity positions under the predicted distribution, and (b) average normalized pixel distance (coordinates normalized by image height and width) of the ground truth from the most likely predicted entity location. While NLL captures both location and scale, pixel distance only measures location accuracy. We report metrics on unseen test descriptions using ground truth locations and appearances for previous entities in the partial video.\\

\input{tables/stage1_ablation.tex}

\noindent \textbf{Feature Ablation.} The ablation study in Table~\ref{table:stage1_ablation} shows that the layout composer benefits from each of the 3 input features -- text, scene context (partial video), and 2D coordinate grid. The significant drop in NLL without text features indicates the importance of entity identity, especially in predicting scale. The lack of spatial awareness in convolutional feature maps without the 2D coordinate grid causes pixel distance to approximately double. The performance drop on removing scene context is indicative of the relevance of knowing \textit{what} entities are \textit{where} in the scene in predicting the location of next entity. Finally, replacing vanilla convolutions by dilated convolutions increases the spatial receptive field without increasing the number of parameters improves performance, which corroborates the usefulness of scene context in layout prediction.\\


\vspace{-2.0em}
\subsection{Entity Retriever Evaluation.}

\vspace{-0.5em}
\noindent \textbf{Training.} We use the Adam optimizer (learning rate=$0.001$, decay factor=$0.5$ every $10$ epochs) and a batch size of 30.\\

\noindent \textbf{Metrics.} To evaluate semantic fidelity of retrieved entities to the query caption, we measure noun, adjective, and verb recalls ($@1$ and $@10$) averaged across entities in the test set. The captions are automatically parsed to identify nouns, adjectives and verbs associated with each entity both in the query captions and target database (using GT database captions for evaluation only). Note that captions often contain limited adjective and verb information. For example, a \emph{red hat} in the video may only be referred to as a \emph{hat} in the caption, and \emph{Fred standing and talking} may be described as \emph{Fred is talking}. We also do not take synonyms (\emph{talking}-\emph{speaking}) and hypernyms (\emph{person}-\emph{woman}) into account. Thus the proposed metric underestimates performance of the entity retriever.\\

\noindent \textbf{Feature Ablation.} Table~\ref{table:stage2_feature_ablation} shows that text and location features are critical to noun, adjective and verb recall. Scene context only marginally affects noun recall but causes significant drop in adjective and verb recalls.\\

\input{tables/stage2_feature_ablation.tex}
\input{tables/stage2_loss_ablation.tex}

\input{tables/train_vs_test.tex}

\noindent \textbf{Effect of Auxiliary Loss.} Table~\ref{table:stage2_loss_ablation} shows that triplet loss alone does significantly worse than in combination with auxiliary classification loss. Adding the auxiliary classification loss on either query or target embeddings improves over triplet only but is worse than using all three. Interestingly, using both auxiliary losses outperforms triplet loss with a single auxiliary loss (and triplet only) on adjective and verb recall. This strongly suggests the benefits of multi-task training in entity retrieval.\\

\vspace{-0.5em}
\noindent \textbf{Background retriever.} Similar to the entity recall evaluation, we computed a top-1 background recall of $57.5$ for \model.\\

\vspace{-0.5em}
\noindent \textbf{Generalization to unseen videos.} A key advantage of the embedding based text to entity video retrieval approach over text only methods is that the embedding approach can use any unseen video databases without any text annotations, potentially in entirely new domains (eg. learning from synthetic video caption datasets and applying the knowledge to generate real videos). However, this requires a model that generalizes well to unseen captions as well as unseen videos. In Table~\ref{table:train_vs_test} we compare entity recall when using the train set (seen) videos as the target database vs using the test set (unseen) video as the target database. \\

\vspace{-0.5em}
\noindent \textbf{OHEM vs All Mini-Batch Triplets.} We experimented with online hard example mining (OHEM) where negative samples that most violate triplet constraints are used in the loss. All triplets achieved similar or higher top-1 noun, adjective and verb recall than OHEM when querying against seen videos ($1.8$,$75.3$,$8.5\%$ relative gain) and unseen videos ($1.7,42.8,-5.0\%$ relative gain).  \\

\noindent \textbf{Modelling Whole Video vs Entities.} A key motivation to composing a scene from entities is the combinatorial nature of complex scenes. To illustrate this point we compare \model\ to a text-to-text based whole video retrieval baseline. For a given test caption, we return a video in the database whose caption has the highest BLEU-1 score. This approach performs much worse than our model except on verb recall (BLEU: $49.57,5.18,26.64$; Ours: $62.3,21.7,16.0$). This indicates that novel captions often do not find a match in the target database with all entities and their attributes present in the same video. However, it is more likely that each entity and attribute combination appears in some video in the database. Note that text-to-text matching also prevents extension to unseen video databases without text annotations.\\

\vspace{-2.0em}
\subsection{Human Evaluation}
\vspace{-0.5em}
\noindent \textbf{Metrics.} In addition to the automated recall metrics which capture semantic fidelity of the generated videos to the captions, we run a human evaluation study to estimate the \emph{compositional consistency} of entities in the scene (given the description) and the overall \emph{visual quality} (independent of the description). The consistency metric requires humans to rate each entity in the video on a $0$-$4$ scale on three aspects: (a) \emph{position} in the scene, (b) \emph{size} relative to other entities or the background, and (c) appearance and consistency of described \emph{interactions} with other entities in the scene. The visual quality metric measures the aesthetic and realism of the generated scenes on a $0$-$4$ scale along three axes: (a) \emph{foreground quality}, (b) \emph{background quality}, and (c) \emph{sharpness}. See supplementary material for the design of these experiments.\\

\input{tables/full_system_evaluation.tex}
\vspace{-0.5em}
\noindent \textbf{Modelling Pixels vs Retrieval.} We experimented extensively with text conditioned whole video generation using models with and without adversarial losses and obtained poor results. Since generative models tend to work better on images with single entities, we swapped out the target embedding network in the entity retriever by a generator. Given the query embedding at each of the $F$ time steps, the generator produces an appearance image and a segmentation mask. The model is trained using an $L1$ loss between the masked appearance image and the masked ground truth image, and an $L1$ loss between the generated and ground truth masks. See supplementary material for more details. This baseline produced blurry results with recognizable colors and shapes for most common characters like \emph{Fred, Wilma, Barney}, and \emph{Betty} at best. We also tried GAN and VAE based approaches and got only slightly less blur. Table~\ref{table:full_system_evaluation} shows that this model performs poorly on the Visual Quality metric compared to \model. Moreover, since the visual quality of the generated previous entities affects the performance of the layout composer, this also translates into poor ratings on the composition consistency metric. Since the semantic fidelity metrics can not be computed for this pixel generation approach, we ran a human evaluation to compare this model to ours. Humans were asked to mark nouns, adjectives and verbs in the sentence missing in the generated video. \model\ significantly outperformed the pixel generation approach on noun, adjective, and verb recall (\model\: $61.0,54.5,67.8$, L1: $37.8,45.9,48.1$).\\

\vspace{-1.0em}
\noindent \textbf{Joint vs Independent Modelling of Layout.} We compare \model\ to a model that uses the same entity retriever but with ground truth (GT) positions. Using GT positions performed worse than \model\ (GT:$62.2,18.1,12.4$; Full:$62.3,21.7,16.0$ Recall@1). This is also reflected in the composition consistency metric (GT:$1.69,1.69,1.34$; Full:$1.78,1.89,1.46$). This emphasizes the need to model layout composition and entity retrieval jointly. When using GT layouts, the retrieval gets conditioned on the layout but not vice versa.

\input{figures/qualitative.tex}

%% file: tables/stage1_ablation.tex
\setlength{\tabcolsep}{5pt}
\begin{table}[ht]
\begin{center}
\vspace{-0.5em}
\caption{\textbf{Layout Composer Analysis.} Evaluation of our model (last row) and ablations on test set. First row provides theoretically computed values assuming a uniform location distribution while making no assumptions about the scale distribution.}
\vspace{-0.5em}
\label{table:stage1_ablation}
\begin{tabular}{cccccc}
\toprule
\thead{Text} & \textbf{\thead{Scene Context}} & \thead{2D Coord. Grid} & \thead{Dil. Conv} & \thead{NLL} & \thead{Pixel Dist.}\\
\hline
\multicolumn{4}{c}{Uniform Distribution}          & \textgreater 9.704 & \textgreater 0.382 \\
\hline
\xmark       & \cmark & \cmark & \cmark    & 9.845    & 0.180                   \\
\cmark &    \xmark        & \cmark & \cmark    & 8.167    & 0.185              \\
\cmark & \cmark &    \xmark        & \cmark    & 8.250    & 0.287              \\
\cmark & \cmark & \cmark &      \xmark         & 7.780    & 0.156              \\ 
\hline
\cmark & \cmark & \cmark & \cmark    & \textbf{7.636}    & \textbf{0.148}       \\       
\bottomrule

\end{tabular}
\end{center}
\vspace{-2.75em}
\end{table}
\setlength{\tabcolsep}{1.4pt}

%% file: tables/stage2_feature_ablation.tex
\setlength{\tabcolsep}{4pt}
\begin{table}[htbp]
\begin{center}
\vspace{-0.5em}
\caption{\textbf{Entity retriever feature ablation.} Top-1 and top-10 recalls of our model (last row) and ablations while generating videos for unseen test captions.}
\vspace{-0.5em}
\label{table:stage2_feature_ablation}
\begin{tabular}{ccccccccc}
\toprule
\multicolumn{3}{c}{\textbf{Query Features}} & \multicolumn{3}{c}{\textbf{Recall@1}}   & \multicolumn{3}{c}{\textbf{Recall@10}}              \\ \cmidrule(l){1-3} \cmidrule(l){4-6} \cmidrule(l){7-9}
\textbf{Text}   &   \textbf{Context}  &   \textbf{Location} & \textbf{Noun}  & \textbf{Adj.}   & \textbf{Verb}      & \textbf{Noun}  & \textbf{Adj.}   & \textbf{Verb}  \\
\hline
\xmark           & \cmark & \cmark & 24.88 & 3.04  & 9.48  & 55.22 & 19.39 & 37.18 \\
\cmark &      \xmark      & \cmark & 60.54 & 9.5   & 11.2  & \textbf{77.71} & 39.92 & 43.58 \\
\cmark & \cmark &    \xmark        & 56.14 & 8.56  & 11.34 & 73.03 & 39.35 & 41.48 \\ \hline
\cmark & \cmark & \cmark & \textbf{61.19} & \textbf{12.36} & \textbf{14.77} & 75.98 & \textbf{47.72} & \textbf{46.86} \\ \bottomrule
\end{tabular}
\end{center}
\vspace{-1.0em}
\end{table}
\setlength{\tabcolsep}{1.4pt}

%% file: tables/stage2_loss_ablation.tex
\setlength{\tabcolsep}{4pt}
\begin{table}[h]
\begin{center}
\vspace{-0.5em}
\caption{\textbf{Entity retriever loss ablation.} Top-1 and top-10 recalls of our model (last row) and ablations while generating videos for unseen test captions.}
\vspace{-0.5em}
\label{table:stage2_loss_ablation}
\begin{tabular}{ccccccccc}
\toprule
& \multicolumn{2}{c}{\textbf{Auxiliary Loss}} & \multicolumn{3}{c}{\textbf{Recall@1}}   & \multicolumn{3}{c}{\textbf{Recall@10}}              \\ \cmidrule(l){2-3} \cmidrule(l){4-6} \cmidrule(l){7-9}
\textbf{Triplet}   &   \textbf{Query}  &   \textbf{Target} & \textbf{Noun}  & \textbf{Adj.}   & \textbf{Verb}      & \textbf{Noun}  & \textbf{Adj.}   & \textbf{Verb}  \\
\hline
\xmark           & \cmark & \cmark    & 35.75 & 7.79  & 8.83  & 63.62 & 43.35 & 33.12 \\
\cmark &    \xmark        & \cmark    & 51.68 & 3.8   & 8.66  & 67.86 & 25.28 & 39.46 \\
\cmark & \cmark &    \xmark           & 50.54 & 4.94  & 9.94  & 66.36 & 28.52 & 39.5  \\
\cmark &   \xmark         &      \xmark         & 48.59 & 3.04  & 9.34  & 65.64 & 20.15 & 37.95 \\ \hline
\cmark & \cmark & \cmark & \textbf{61.19} & \textbf{12.36} & \textbf{14.77} & \textbf{75.98} & \textbf{47.72} & \textbf{46.86} \\ \bottomrule
\end{tabular}
\end{center}
\vspace{-1.0em}
\end{table}
\setlength{\tabcolsep}{1.4pt}

%% file: tables/train_vs_test.tex
\setlength{\tabcolsep}{3pt}
\begin{table}[!h]
\begin{center}
\vspace{-0.5em}
\caption{\textbf{Generalization to Unseen Database Videos.} Retrieval results for \model\ when queried against seen videos vs unseen videos.}
\vspace{-0.5em}
\label{table:train_vs_test}
\begin{tabular}{ccccccc}
\toprule
\multirow{2}{*}{\textbf{Video Database}} & \multicolumn{3}{c}{\textbf{Recall@1}}   & \multicolumn{3}{c}{\textbf{Recall@10}}              \\ \cmidrule(l){2-4} \cmidrule(l){5-7}
 & \textbf{Noun}  & \textbf{Adj.}   & \textbf{Verb}      & \textbf{Noun}  & \textbf{Adj.}   & \textbf{Verb}  \\
\hline
Seen (Train) & 61.19 & 12.36 & 14.77 & 75.98 & 47.72 & 46.86 \\
Unseen (Test) & 50.52 & 11.98 & 10.4  & 69.1  & 41.25 & 42.57 \\ \bottomrule
\end{tabular}
\end{center}
\vspace{-2.75em}
\end{table}
\setlength{\tabcolsep}{1.4pt}

%% file: tables/full_system_evaluation.tex
\setlength{\tabcolsep}{2pt}
\begin{table}[t]
\begin{center}
\vspace{-0.5em}
\caption{\textbf{Human evaluation} to estimate consistency and quality of generated videos.}
\vspace{-0.5em}
\label{table:full_system_evaluation}
\begin{tabular}{ccccccc}
\toprule
 &  \multicolumn{3}{c}{\textbf{Composition Consistency}} & \multicolumn{3}{c}{\textbf{Visual Quality}}\\ \cmidrule(l){2-4} \cmidrule(l){5-7}
 &  \textbf{Position} & \textbf{Rel. Size} & \textbf{Interact.} & \textbf{FG} & \textbf{BG} & \textbf{Sharpness}\\
\hline
Pixel Generation L1 & 0.69 & 0.65 & 0.55 & 0.96 & 1.44 & 1.07\\
Ours (GT Position) & 1.69 & 1.69 & 1.34 & 1.49 & 1.65 & \textbf{2.16}\\
Ours & \textbf{1.78} & \textbf{1.86} & \textbf{1.46} & \textbf{1.98} & \textbf{1.95} & 1.82\\
\bottomrule
\end{tabular}
\vspace{-2.5em}
\end{center}
\end{table}
\setlength{\tabcolsep}{1.4pt}

%% file: figures/qualitative.tex
\begin{figure}[t]
\centering
\vspace{-0.0em}
\includegraphics[width=\textwidth]{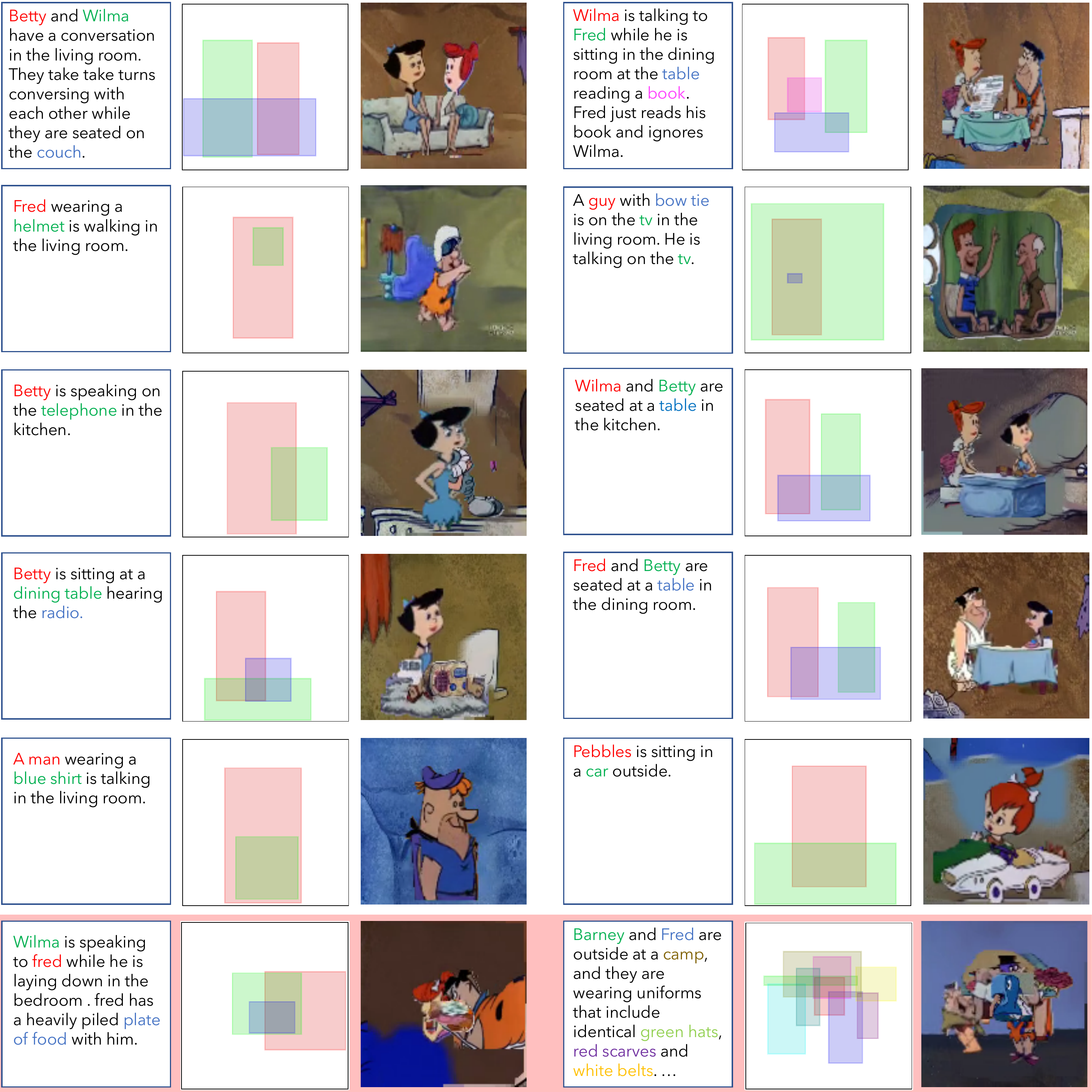}
\vspace{-2.0em}
\caption{\textbf{Qualitative results} for \model. Last row shows failures of the layout composer (left) and the entire system (right). See \url{https://youtu.be/688Vv86n0z8} for video examples, failure cases, and visualization of predicted location and scale distributions.}
\vspace{-2.0em}
\label{fig:qualitative}
\end{figure}

%% file: sections/appendix.tex
\title{Supplementary Material}
\author{}
\institute{}
\maketitle

\vspace{-2.0em}
\section*{Overview}
The supplementary material in this PDF is organized as follows:
\begin{itemize}
    \item Section A: Derivation of Uniform distribution inequalities in Table 1
    \item Section B: Details on the \dataset\ dataset including dataset collection and dataset diversity
    \item Section C: Details on the design of human evaluation tasks with Amazon Mechanical Turk interfaces
\end{itemize}

In addition, more qualitative results (in video form) can be found in the video included with this supplementary material.

\section{Derivation of Uniform distribution inequalities}

The first row of Table 1 in the submitted paper provides theoretically computed values with a uniform location distribution and no assumptions about the scale distribution. Here we provide derivations for the same.

\subsection{Negative Log Likelihood}
\begin{align}
P(x,y,w,h) &= P(x,y)P(w,h|x,y) \\
&\leq P(x,y) \\
\implies \log{P(x,y,w,h)} & \leq \log{P(x,y)} \; (\text{Since, } P(w,h|x,y) \leq 1)\\
&= \log{\frac{1}{128\times 128}} \; (\text{For a } 128\times 128 \text{ image})\\
\implies -\log{P(x,y,w,h)} &\geq 9.704 
\end{align}

\noindent \textbf{Typo:} Please note a small typo in Table 1 in the main submission. The negative log likelihood for uniform distribution is $\geq 9.704$ instead of $<9.704$.

\subsection{Pixel Distance}
Let $(x_t,y_t)$ be the target (ground truth) location. The expected normalized distance (normalizing location coordinates to $[0,1]$ range) from ground truth location given the predicted distribution is given by
\begin{align}
& E[\|(x-x_t)/128,(y-y_t)/128\|_2] \\
&= \int_0^{128}\int_0^{128} \|x-x_t,y-y_t\|_2 P(x,y) dxdy \\
&= \int_0^{128}\int_0^{128} \|(x-x_t)/128,(y-y_t)/128\|_2 \; \frac{1}{128\times 128}\; dxdy
\end{align}
Note that the best case scenario is $x_t=128/2$ and $y_t=128/2$ (the target lies at the center of the image). Hence,
\begin{align}
& E[\|(x-x_t)/128,(y-y_t)/128\|_2] \\
&\geq \int_0^{128}\int_0^{128} \|x/128-0.5,y/128-0.5\|_2 \; \frac{1}{128\times 128} \; dxdy \\
&= \int_0^{1}\int_0^{1} \|x'-0.5,y'-0.5\|_2 \; dx'dy' \; (\text{Substituting } x'=x/128, y' = y/128)\\
&= 0.382 \; (\text{Solved using Wolfram Alpha})
\end{align}

\section{\dataset\ Dataset}

\subsection{\dataset\ Dataset Construction}

\noindent \textbf{Clip Generation.} \dataset\ clips are roughly three seconds (75 frames) in duration, a length chosen to capture a small number of discrete actions while limiting the number of sentences needed to describe them. Source videos for the \dataset\ begin as episode-length videos with no existing subdivisions. In order to assure that our clips don’t span scene and shot changes, we first locate these by detecting abrupt frame-to-frame changes and subdivide between them.\\

\noindent \textbf{Clip Annotation.} An outline of our annotation process is shown in Figure~\ref{fig:dataset_collection}. Raw clips enter the first stage of our annotation pipeline, where crowdworkers identify and localize characters by concurrently labelling them and providing their bounding boxes in three keyframes. For a small number of recurring characters (e.g. Fred, Wilma), we allow workers to select from predefined labels, while for others they write a brief description (e.g. policeman, old man in red shirt). Clips containing between 1-4 characters are passed to the next stage, where workers provide a 1-2 word description of a clip's setting (e.g. living room, park). In the third stage of the pipeline, crowdworkers write a 1-4 sentence description of the clip using the established character and setting labels. A fourth task identifies important objects mentioned in descriptions, which are annotated with bounding boxes in a fifth and final stage of the pipeline.\\ 

\noindent \textbf{Annotation Supplementation.} The dataset also provides tight segmentation masks for characters and objects, as well as clean scene backgrounds. The prohibitive cost of human-annotated masks necessitates an automated approach. First, template-matching is used to track entity positions, with the addition of a penalty term for displacement from the interpolated trajectory for stability. Tracking is run forward and backward, and the resulting trajectories are averaged to produce the final entity trajectory. Once an entities' trajectory is established, A rough segmentation mask is produced by using SLIC (Simple Linear Iterative Clustering)~\cite{Achanta2012Slic} to generate superpixels within a frame, which are then merged hierarchically to produce regions of near uniformity. Regions overlapping an entities’ bounding box are joined to form a rough mask. This mask is used to initialize GrabCut~\cite{Rother2004Grabcut}, which further refines the segmentation. Clean backgrounds are the final component of a clip generated. Foreground characters and objects are excised, and the resulting holes are filled using PatchMatch~\cite{Barnes2009PatchMatch}. For static backgrounds, a single median background frame is used for the entire clip, while individual frames are produced for moving backgrounds.

\input{figures/dataset_collection.tex}

\subsection{Dataset Diversity}

\input{figures/wordclouds.tex}

\dataset\ contains a wide variety of named characters, objects, and vocabulary used in describing the actions and appearances of entities and scenes as seen below. Notes: No stemming or lemmatization is performed prior to computing these statistics. Part of Speech tags were obtained using the Stanford core parser~\cite{manning-EtAl:2014:P14-5}.
\begin{itemize}
    \item Number of unique characters: 3897
    \item Number of unique objects: 2614
    \item Number of unique verbs: 1350
    \item Number of unique settings: 323
\end{itemize}
Figure~\ref{fig:wordclouds} demonstrates a fraction of this diversity, with the most frequent character, objects, verbs and settings rendered in word clouds. 

\section{Design of Human Evaluation Tasks}

To compute \emph{compositional consistency} and \emph{visual quality} (metrics mentioned in Section 5.3 in the submitted paper), we ran a human evaluation study on Amazon Mechanical Turk.

The consistency metric requires humans to rate each entity in the video on a $0$-$4$ scale on three aspects: (a) \emph{position} in the scene, (b) \emph{size} relative to other entities or the background, and (c) appearance and consistency of described \emph{interactions} with other entities in the scene. 

The visual quality metric measures the aesthetic and realism of the generated scenes on a $0$-$4$ scale along three axes: (a) \emph{foreground quality}, (b) \emph{background quality}, and (c) \emph{sharpness}. 

Workers were provided several examples of common defects to calibrate their ratings, in addition to written guidelines. Each video was given to three workers and their ratings averaged. To assure worker consistency between models, tasks were run simultaneously for all models and ablations. Figure~\ref{fig:interface_2} shows the turk interface used for \emph{visual quality} and Figure~\ref{fig:interface_1} shows the mechanical turk interface used for \emph{compositional consistency}.

\input{figures/metric_interface_quality.tex}
\input{figures/metric_interface_consistency.tex}


%% file: figures/dataset_collection.tex
\begin{figure}[!h]
\centering
\includegraphics[width=\textwidth]{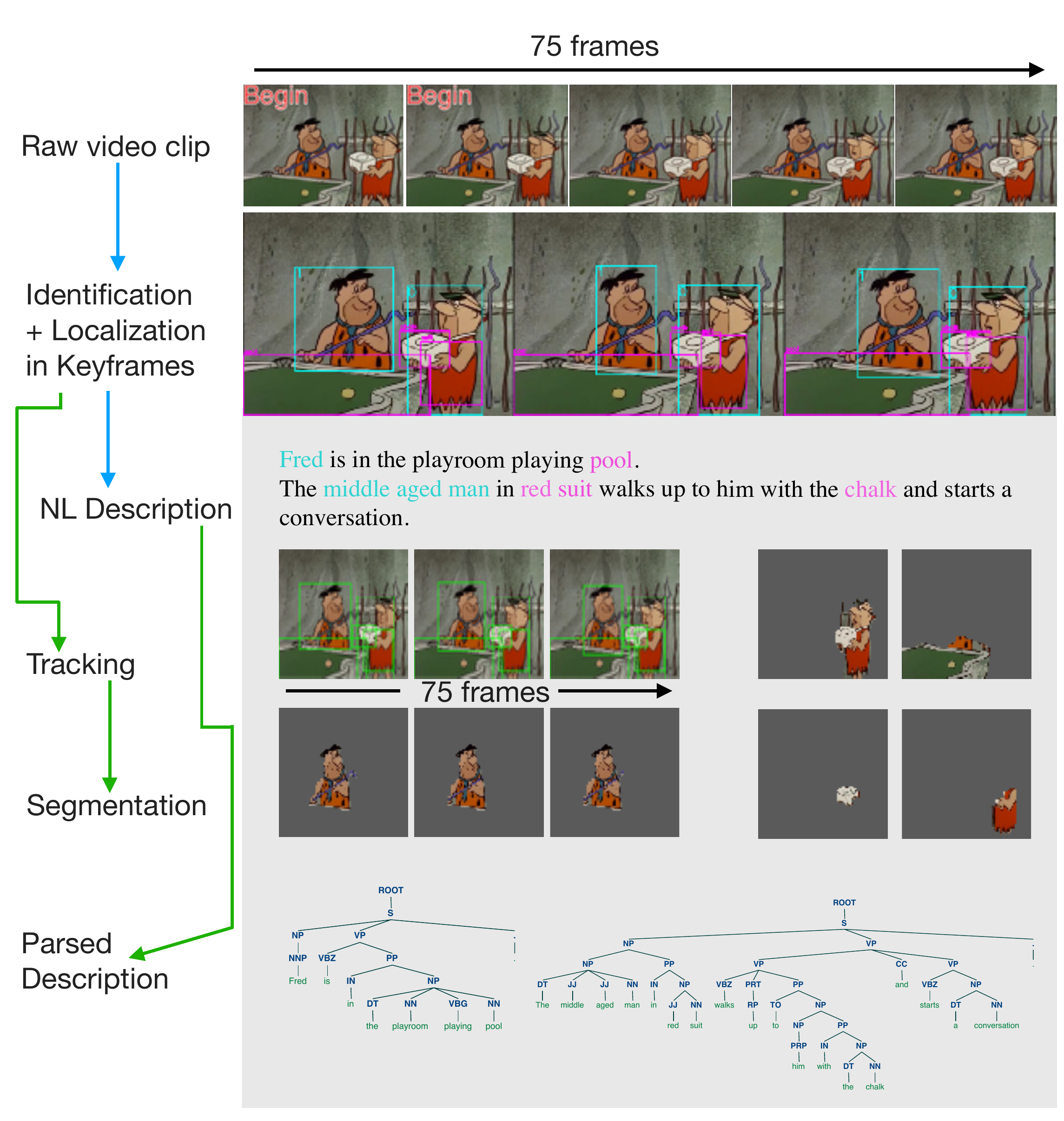}
\caption{An overview of \dataset\ Clip construction. Clip annotations are built up over several stages, each requiring and building on previous stages. Human annotation steps are denoted with a blue arrow and automated data supplementation with a green arrow.}
\label{fig:dataset_collection}
\end{figure}

%% file: figures/wordclouds.tex
\begin{figure}[h]
\centering
\vspace{-2.0em}
\includegraphics[width=\textwidth]{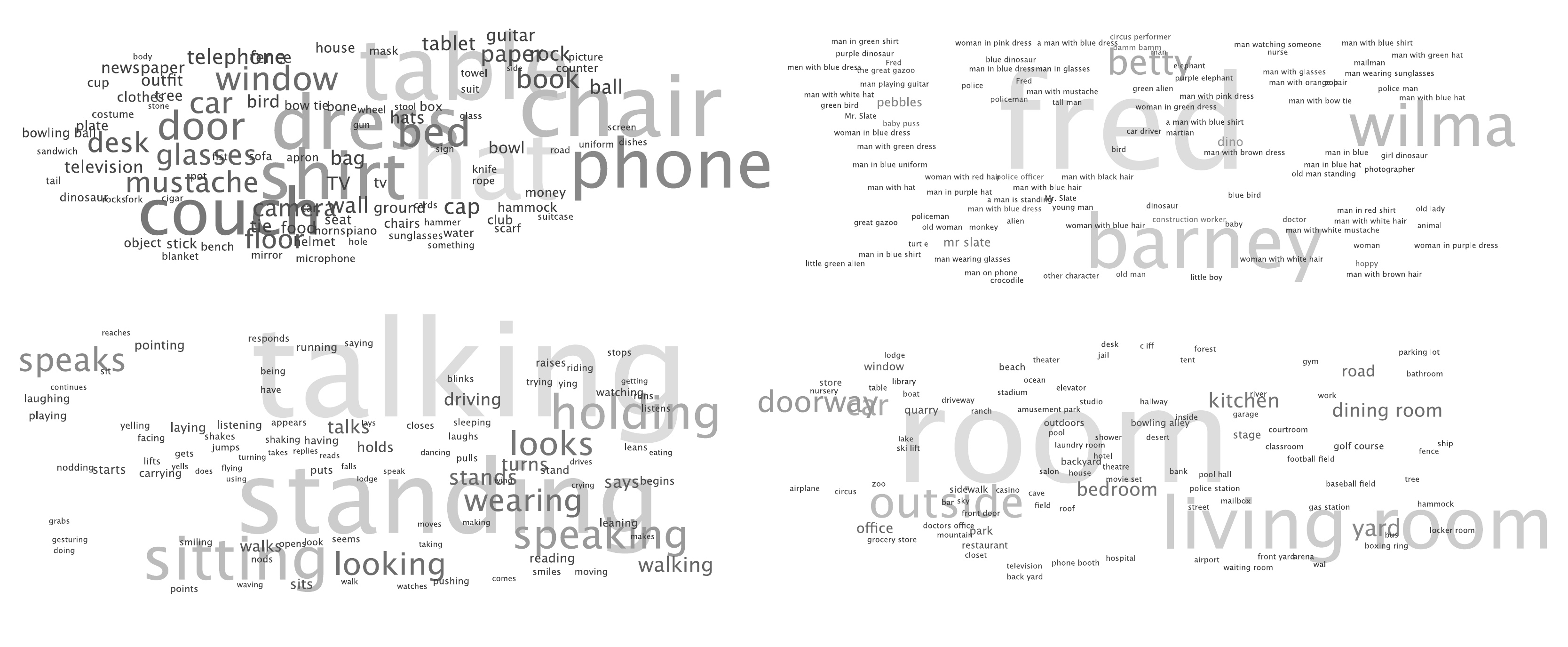}
\caption{Word clouds demonstrating the diversity of the \dataset\ dataset. Each word cloud was created using just the top 100 words within each category, to enable a clearer visualization. (1) Top left: Top 100 Objects (2) Top right: Top 100 Characters (3) Bottom left: Top 100 verbs (4) Bottom right: Top 100 settings.}
\label{fig:wordclouds}
\end{figure}

%% file: figures/metric_interface_quality.tex
\begin{figure}[h]
\centering
\includegraphics[width=\textwidth]{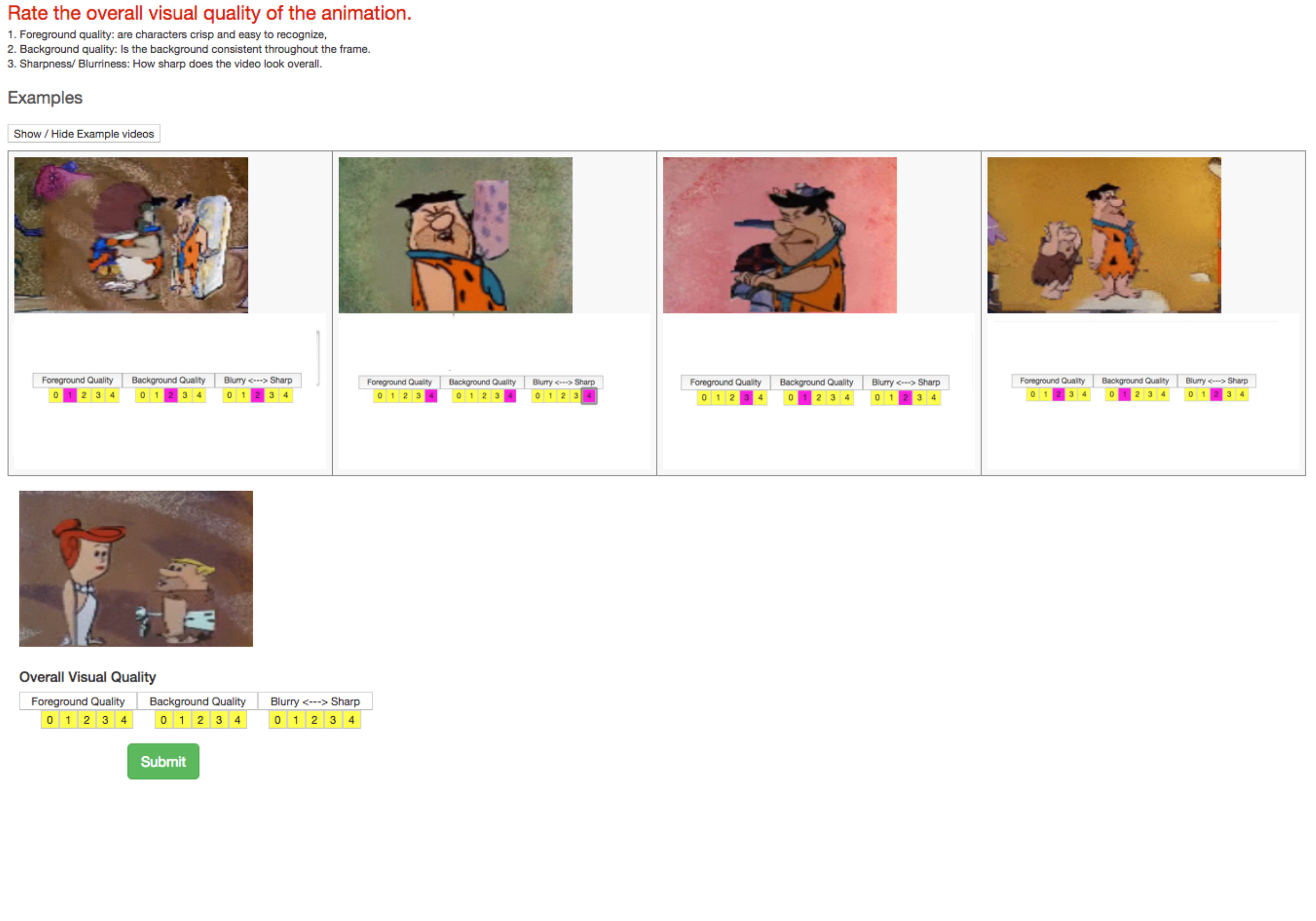}
\vspace{-4em}
\caption{Amazon Mechanical Turk interface design used to collect the \emph{visual quality} metric. This metric is reported in section 5.3 in the submitted paper.}
\vspace{-1.5em}
\label{fig:interface_2}
\end{figure}

%% file: figures/metric_interface_consistency.tex
\begin{figure}[h]
\centering
\includegraphics[width=\textwidth]{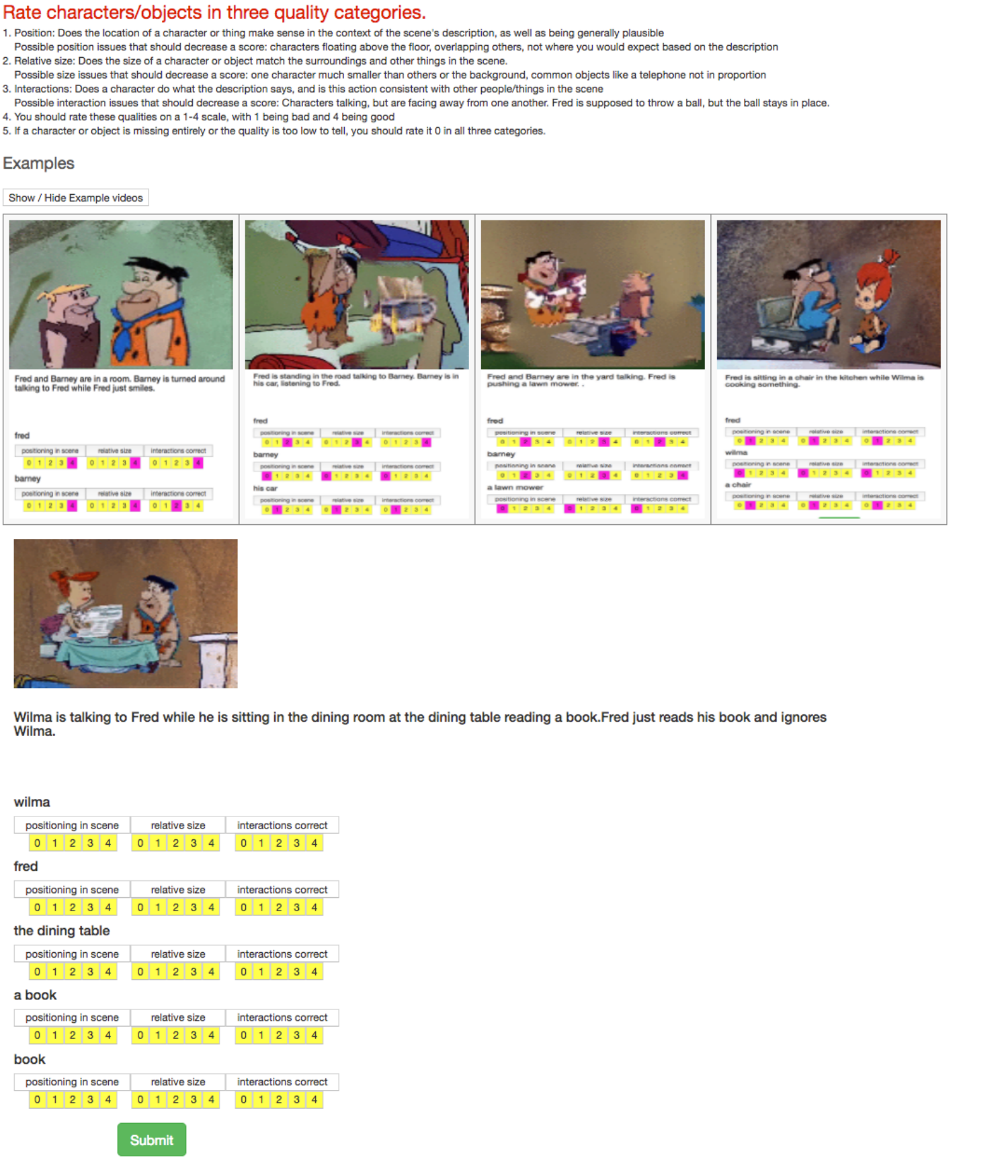}
\caption{Amazon Mechanical Turk interface design used to collect the \emph{compositional consistency} metric. This metric is reported in section 5.3 in the submitted paper.}
\label{fig:interface_1}
\end{figure}